\documentclass[11pt]{article}
\pdfoutput=1 
\usepackage{amsfonts,amsmath,amssymb,amsthm}
\usepackage{dsfont,bm,color}
\usepackage{harvard}
\usepackage{fullpage} 
\usepackage{graphicx}
\citationmode{abbr}
\usepackage[noend]{algpseudocode}
\usepackage{harvard}
\usepackage{times}
\usepackage{textcomp}
\newcommand{\argmin}{\mathop{\mathrm{argmin}}}

\def\R{\mathds{R}}
\def\E{\mathbb{E}}

\def\Var{\mathrm{Var}}

\def\half{\frac{1}{2}}

\def\hbeta{\hat{\beta}}
 \def\cS{{\cal S}}

\usepackage{graphicx,amsmath,amsfonts,amscd,amssymb,bm,epsfig,epsf,url,color}
\usepackage[small,bf]{caption}
\usepackage{subcaption}
\author{Nadine
  Hussami and Robert Tibshirani \date{Department of Electrical Engineering and\\
Departments  of  Health Research and Policy, and Statistics,\\
    Stanford University, Stanford CA.;\\nadinehu@stanford.edu; tibs@stanford.edu}}

\title{A Component Lasso}
\date{November 2013}

\begin{document}

\maketitle
\vspace{-0.3in}

\begin{abstract}
 We propose a new sparse regression method called the {\em component lasso},
based on a simple idea.
 The method uses the connected-components structure of the 
  sample covariance matrix to split the problem into smaller ones. It then applies the lasso
to  each subproblem separately, obtaining a coefficient vector for each one. 
  Finally, it uses non-negative least squares to recombine the different vectors into a single solution. This step is useful in selecting and reweighting components
  that are correlated with the response. Simulated and
  real data examples show that the component lasso can outperform standard regression methods such as the lasso and elastic net, achieving a lower
  mean squared error as well as better support recovery. 
The modular structure also lends itself naturally to parallel computation.
\end{abstract}

\section{Introduction}
\label{sec:intro}

  
Suppose that we have a response vector $y\in\R^n$, a matrix
$X \in \R^{n\times p}$ of predictor variables and the
usual linear regression setup:
\begin{equation}
\label{eq:model}
y = X\beta^* + \sigma\epsilon,
\end{equation}
where $\beta^* \in \R^p$ are unknown coefficients to be estimated,
$\sigma^2>0$ is the noise variance, and
the components of the noise vector $\epsilon \in \R^n$ are i.i.d.
with $\E[\epsilon_i]=0$ and $\Var(\epsilon_i)=1$.
We assume that $y$ has been centered, and the columns of $X$ are centered and scaled,
so that we can omit an intercept  in the model.
The lasso estimator \cite{lasso,bp}, is defined as
\begin{equation}
\label{eqn:lasso}
\hbeta = \argmin_{\beta\in\R^p} \, 
\half\|y-X\beta\|_2^2 + \lambda\|\beta\|_1,
\end{equation}
where $\lambda \geq 0$ is a tuning parameter, controlling the degree
of sparsity in the estimate $\hbeta$.
 
 Variable selection is important in many modern applications, for which the lasso has proven to be successful. However, this method has known limitations in
 certain settings: there is a solution with  at most $n$ non-zero coefficients when $p>n$, and if a group of relevant variables is highly correlated, 
 it tends to include only one in the model. {These conditions occur frequently in real applications, such as genomics, where we often have a large number of
 predictors that can be divided into highly correlated groups. It is therefore of practical interest to overcome these limitations.}
 
 The elastic net ~\cite{enet} can sometimes improve  the performance of the lasso. The elastic net penalty is the weighted sum of the 
 $\ell_{2}$ and $\ell_{1}$ norms of the coefficient vector to be estimated: $P_{\alpha}(\beta)=\frac{(1-\alpha)}{2}||\beta||_{2}^2 + \alpha||\beta||_{1}$. It is
 equivalent to the ridge
 regression penalty when $\alpha = 0$, and to the lasso penalty when $\alpha =1$. The elastic net solves the following problem:
 
\begin{equation}
\label{eq:en}
\hbeta = \argmin_{\beta\in\R^p} \, 
\half\|y-X\beta\|_2^2 + \lambda P_{\alpha}(\beta).
\end{equation}
 
{The elastic net penalty  is strictly convex, by strict convexity of the  $\ell_{2}$ norm. Using this fact, the authors  provide an upper bound on the 
distance between coefficients that correspond to highly correlated predictors.}
 This guarantees the grouping effect of the elastic net. Moreover, the elastic net solution can have  more than $n$ non-zero coefficients,
 even when $p>n$, since it is equivalent to solving the lasso on an augmented dataset.   
 
It is easy to see that the elastic net regularizes the feature covariance matrix
from $X^TX$ to a form  $X^TX+ \frac{1-\alpha}{2}\cdot\lambda I_p$ where
$I_p$ is the $p\times p$ identity matrix. By inflating the diagonal
 it reduces the effective size of the off-diagonal correlations. 
{If the feature covariance matrix is block diagonal, its connected components
 correspond to groups of predictors that are correlated with each other but not with predictors in other groups.} 
Here, we introduce a method adapted to situations where the sample covariance matrix is approximately block diagonal. Our proposed method, {\em the
component lasso}, applies a more severe form of decorrelation than the elastic net to exploit this structure.

 Consider the inverse of the covariance matrix of the predictors. Zeros in this matrix correspond to conditionally independent variables.
Recent  work has focused on estimating 
a sparse version of the inverse covariance by optimizing the $\ell_1$ penalized
 log-likelihood.
 The so-called ``graphical lasso'' algorithm solves the problem by cycling through the variables and fitting a modified lasso regression to each one.
In their ``scout'' procedure, \citeasnoun{WT2009} used the graphical lasso in a penalized regression framework to estimate the inverse covariance 
of $X$. Then they applied a modified form of the lasso to estimate the regression parameters.

 More recently, a connection between the graphical lasso and connected components has been established by \citeasnoun{WFS2011} and \citeasnoun{MH2012}.
Specifically, the connected components in the  estimated inverse covariance matrix
correspond exactly to those obtained from single-linkage clustering 
of the correlation matrix. Clustering the correlated variables before estimating the parameters has been suggested by \citeasnoun{GE2007}
and \citeasnoun{BUH2007}.

\begin{figure}[!h]
  \centering
  \includegraphics[scale=0.7]{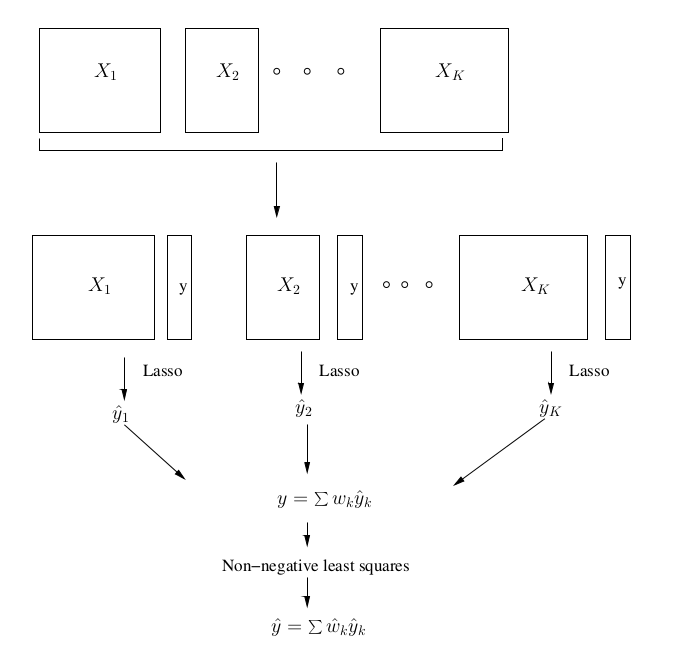}
  \caption{\em The component lasso steps: The predictors are split according to the
estimated connected components of
the sample covariance matrix. The lasso is applied to each subset of predictors to
separately  estimate the coefficients  and to predict the response. Finally, the
different coefficient vectors are combined using a non-negative least
squares fit of $y$ on the $K$ predictions from each component. }
  \label{fig:clgph}
 \end{figure}

In this paper, we propose a new simple idea to make use of the connected components in penalized regression.
The {\em component lasso} 
works by  (a) finding the connected components of the estimated covariance matrix, (b) solving separate
lasso problems for each component, and then (c) combining the componentwise predictions into one final prediction.
We show that this approach can
 improve the accuracy, and interpretability of the lasso and elastic net methods. 
The method  is summarized in Figure \ref{fig:clgph}.

\begin{figure}    
\begin{minipage}[t]{0.45\textwidth}
\includegraphics[width=\linewidth]{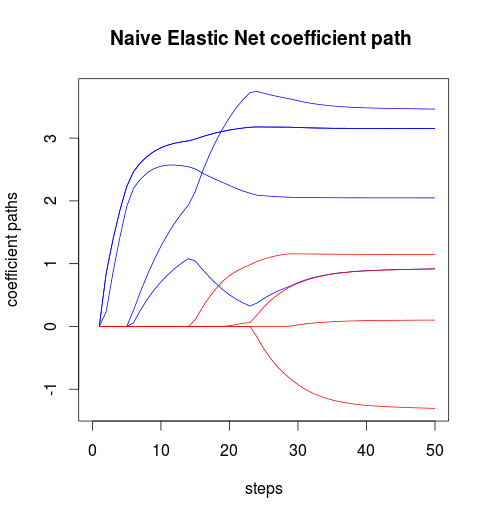}
\end{minipage}
\hspace{\fill}
\begin{minipage}[t]{0.45\textwidth}
\includegraphics[width=\linewidth]{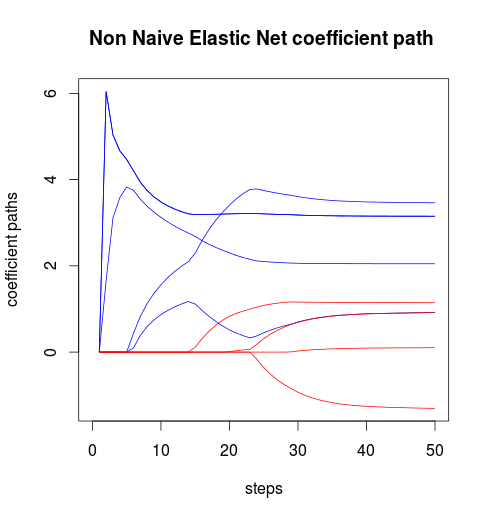}
\end{minipage}

\vspace*{0.5cm} 
\begin{minipage}[t]{0.45\textwidth}
\includegraphics[width=\linewidth]{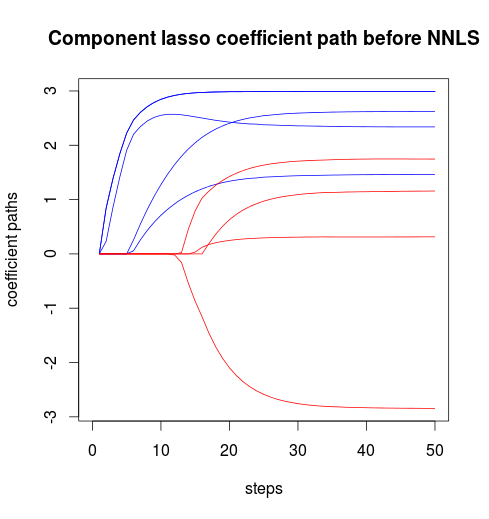}
\end{minipage}
\hspace{\fill}
\begin{minipage}[t]{0.45\textwidth}
\includegraphics[width=\linewidth]{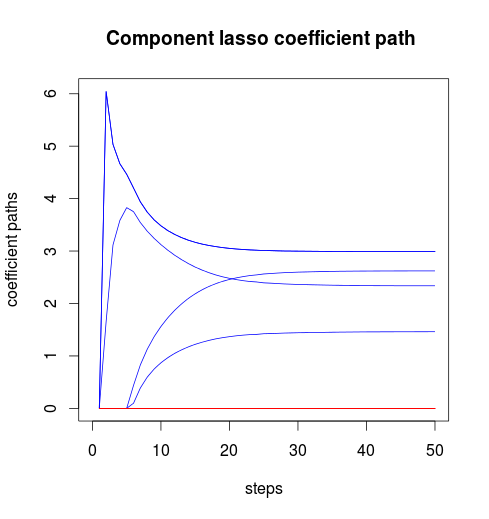}
\end{minipage}
\caption{\em Coefficient paths for : the naive elastic net (top left), the non naive elastic net (top right), the component lasso before non-negative
least squares (bottom left), and the component lasso (bottom right).
The signal variables are shown in blue, while the non-signal variables are in red.}
\label{fig:paths}
\end{figure}

 The following example motivates the remainder of the paper. Consider eight predictors, and let the corresponding covariance matrix be block diagonal with
 two blocks. Suppose that the predictors corresponding to the first block, or equivalently component,
 are all signal variables. The second component only contains noise variables. Figure~\ref{fig:paths} shows the coefficient paths for the naive
 and non naive elastic net, and
 the component lasso before and after the non-negative least squares (NNLS) recombination step when the sample covariance is split into two blocks.
 The paths are plotted for all values of the tuning parameter $\lambda$. 

The example shows the role that NNLS plays in selecting the relevant component which contains the signal variables (in blue)
and reducing the coefficients of the noise variables (in red) in the second component to zero. This
illustrates the possible improvements that can be achieved by finding the block-diagonal structure of the sample covariance matrix, as compared 
to standard methods.

The remainder of the paper is organized as follows. We explain our algorithm in Section \ref{sec:comp_lass}. 
 Section \ref{sec:ex} includes simulated 
and real data results.
Section \ref{sec:comp} focuses on the computational complexity of the component lasso, and
presents ideas for making it more efficient.
We conclude the paper with a short discussion in Section \ref{sec:discussion}, including possible extensions to generalized linear models.

\section{The Component Lasso}
\label{sec:comp_lass}

\subsection{The main idea}

 The lasso minimizes the  $\ell_1$ penalized criterion
(\ref{eqn:lasso})
 whose corresponding subgradient equation is
\begin{equation}
  X^TX\beta - X^T y + \lambda\cdot \text{sign}(\beta) = 0,
\label{eqn:subg}
\end{equation}
where $\text{sign}(\beta)$ is  a vector with components $s_j=\text{sign} (\beta_j)$ if $\beta_j\neq 0$ and
$s_j \in [-1,1]$ if $\beta_j= 0$.

 The solution to the lasso can be written as
\begin{equation}
 \hat{\beta} = (X^TX)^{-}(X^Ty-\lambda\cdot  \text{sign}(\hat{\beta}))
\label{eqn:lassosol}
\end{equation}
where $(X^TX)^{-}$ represents a generalized inverse of $X^TX$.

 
   Let $\Sigma=\text{cov}(X)$. We propose replacing $(X^TX)^{-}$ by a block diagonal estimate $n^{-1}\hat\Theta \approx n^{-1}\Sigma^{-1} $, the blocks 
   of $\hat\Theta$ being
   the (estimated) connected components. Finding $K$ connected components splits the subgradient equation into $K$ separate equations:
   \begin{equation}
   \label{eqn:subg2}
    X^T_k X_k \beta_k - X_k^T y + \lambda\cdot \text{sign}(\beta_k ) = 0
   \end{equation}
   for $k = 1, 2,\dots K$, where $X_k$ is a subset of $X$ containing the observations of the predictors in the $k$th component, and $\beta_k$ contains the 
   corresponding coefficients.
   
   Each subproblem can be solved individually using a standard lasso or elastic net algorithm. The resultant coefficients $\beta_k$
   are then combined into a solution to the original problem. The use of the block-diagonal covariance matrix
   creates a substantial bias in the coefficient estimates, so the combination step is quite important.
   We scale the componentwise solution vectors
   $\hat\beta_1, \hat\beta_2, \ldots \hat\beta_K$ using a non-negative least squares refitting of $y$ on $\{\hat{y}_k=X_k\hat\beta_k\}, k=1, \dots, K$.
   The non-negativity constraint seems natural since each componentwise predictor
   should have positive correlation with the outcome.
   
   The component lasso objective function,  corresponding to a block diagonal estimate of the sample covariance with connected components
   $C_1,\dots,C_K$ is:
   \begin{eqnarray}
J(\beta, c)= \sum_ {k=1}^K \sum_{i=1}^n  \left[\frac{1}{2}\left(y_i-c_k\sum_{j\in C_k} X_{ij}\beta_j \right)^2+\lambda\left(\sum_{j \in C_k} \alpha|\beta_j| + 
    \frac{(1-\alpha)}{2}||\beta_j||_2^2 \right) \; \right]
\label{eqn:obj}
\end{eqnarray}
 subject to $c_k\geq 0 \; \forall k$.
Our algorithm (detailed below) sets $c_k=1 \; \forall k$, optimizes over $\beta$, and then optimizes over $c$.

 Consider an extreme case where the sample covariance matrix happens to be block diagonal with $K$ connected components. This occurs when predictors in
 different correlated groups are orthogonal to each other. The subgradient equation of the lasso splits naturally into separate systems of equations
 as in equation (\ref{eqn:subg2}) for the component lasso. The lasso coefficients will be identical to the component lasso coefficients before the NNLS
 step, which reweights the predictors corresponding to each component.

 
 This can be easily extended to the elastic net. Let $\lambda'=\frac{\lambda(1-\alpha)}{2}$ be the tuning parameter corresponding to the $\ell_2$ penalty. 
 The naive elastic net problem can be written as a lasso problem on an augmented data set $(y^*,X^*)$, where $y^*=(y,0)$ is now an $n+p$ vector and 
 $$
X^*=(1+\lambda')^{-1/2}\left[
\begin{array}{c}
X \\
\sqrt {\lambda'} I
\end{array}\right]
.$$ 
The sample covariance matrix corresponding to the augmented observations $X^*$
$$(1+\lambda')^{-1}(X^TX + \lambda'I)$$
is clearly block diagonal when the predictors in different components are orthogonal.

Therefore, the subgradient equation of the elastic net splits as well, and the elastic net coefficients will be identical to those of the component
lasso before the NNLS step. In this case, the model chosen by the component lasso will involve splitting the predictors into components if the NNLS 
reweighting is useful in minimizing the validation MSE.


\subsection{Details of connected component estimation via the graphical lasso}
Given an observed covariance $S={X^TX/n}$ with $X \sim \mathcal{N}(0,\Sigma)$,
the graphical lasso estimates $\Theta=\Sigma^{-1}$ by maximizing the penalized log-likelihood 
   \begin{eqnarray}
\ell(\Theta)=\log{\rm det} \Theta-{\rm tr}(S\Theta)-\tau ||\Theta||_1
\end{eqnarray}
over all non-negative definite matrices $\Theta$ . 
The KKT conditions for this problem are
  \begin{eqnarray}
\Theta^{-1}=S-\tau \Gamma(\Theta)=0
\end{eqnarray}
where $\Gamma(\Theta)$ is a matrix of componentwise subgradients $\text{sign}(\Theta_{ij})$.
If $C_1, C_2 \ldots C_K$ are a partition of $1,2,\ldots p$, then  \citeasnoun{WFS2011} and \citeasnoun{MH2012} show that the corresponding
arrangement of $\hat\Theta(\tau)$ is block diagonal if and only if 
$S_{ii'} \leq \tau$ for all $i \in C_k, i' \in C_{k'}, k\neq k'$.
This means that soft-thresholding of $S$ at level $\tau$ into its connected components yields the 
connected components of $\hat\Theta(\tau)$.

Furthermore, there is an interesting connection
to hierarchical clustering. Specifically
 the connected components correspond to the subtrees from 
when we 
apply single linkage agglomerative clustering to $S$ and then cut the dendrogram at level $\tau$
\cite{TWS2013}.
Single linkage clustering is sometimes not very attractive in practice, 
since it can produce long and stringy clusters and hence components of very unequal size.
However, these same authors show that under regularity conditions on $S$, application of average or complete
linkage agglomerative clustering also consistently estimates the 
connected components.
Hence we are free to use average, single or complete linkage  clustering; we use average
linkage in the examples of this paper.

\subsection{Summary of the component lasso algorithm}

\begin{enumerate}
\item Apply  average, single or complete linkage  clustering to $S=X^TX/n$ and cut the dendrogram
at level $\tau$ to produce components $C_1, C_2, \ldots C_K$.
\item For each component $k=1,2,\ldots K$ and fixed elastic net parameter $\alpha$,
compute a path of elastic net solutions $\hat \beta_{k,\alpha,\tau}(\lambda)$ over a grid of $\lambda$ values.
Let $\hat y_{k,\alpha,\tau}(\lambda)$ be the predicted values from the $k$th fit.
\item  Compute the non-negative least squares (NNLS) fit of $y$ on
$\{\hat y_{1,\alpha,\tau}(\lambda), \hat y_{2,\alpha,\tau}(\lambda), \ldots \hat y_{K,\alpha,\tau}(\lambda)\}$,
yielding weights $\{\hat c_1, \hat c_2, \ldots \hat c_K\}$.
Finally, form the overall estimate $\hat\beta_{\alpha,\tau}(\lambda)  
=\sum_{k=1}^K \hat c_k \hat \beta_{k,\alpha,\tau}(\lambda)$.
\item Estimate optimal values of $\tau, \alpha$ and $\lambda$ by cross-validation.
\end{enumerate}
    
{\bf Remark A}.
The above procedure
partially optimizes the bi-convex objective function (\ref{eqn:obj}) in two stages: it sets $c_k=1 \;\forall k$, optimizes
over $\beta$ and then optimizes over the  $c_k$ with $\hat\beta$ fixed.
Of course one could iterate these steps in the hopes of obtaining at least a local
optimum of the objective function.  But we have found that the simple two-step approach
works well in practice and is more efficient computationally.

{\bf Remark B}. The bias induced by setting blocks of the covariance matrix to zero
can be seen in a simple example.
Let $A$ be a block diagonal matrix with blocks $A_1, A_2$
and let the covariance of the features be $S=A+ \rho ee^T$
where $e$ is a $p$-vector of ones.
Assume that $A_1, A_2$ are positive definite.
Then by the Sherman-Morrison-Woodbury formula
\begin{eqnarray}
S^{-1}=A^{-1} -\frac{\rho A^{-1}ee^TA^{-1}}{1+\rho e^TA^{-1}e}
\label{eqn:smw}
\end{eqnarray}
The coefficients for
the full least squares fit   are $S^{-1}X^Ty$;
if instead we set to zero the covariance elements outside of the
blocks $A_1, A_2$, the estimates become $A_j^{-1} x_j y$ for $j=1,2$.
The second term in (\ref{eqn:smw}) represents the bias in using
$A^{-1}$ in place of $S^{-1}$, and is generally larger as $\rho$ increases.

\section{Examples}
\label{sec:ex}

\subsection{Simulated examples}

In this section, we study the performance of the component lasso in several simulated examples. The results show that the component lasso 
can achieve a lower MSE as well as better support recovery in certain settings when compared to common regression and variable selection methods. We report 
the test error, the false positive rate and false negative rate of the following methods: the lasso, a rescaled lasso,
the lasso-OLS hybrid, ridge regression, and the naive and non-naive elastic net. 
{The non-naive elastic net does not correspond to rescaling the 
naive elastic net solution as suggested in the elastic net paper. Instead, we do a least squares fit of the response $y$ on the response that is predicted
using the coefficients estimated by the naive elastic net.} The error is computed as $(\beta-\hat\beta)^T S(\beta-\hat\beta)$ where S is
the observed covariance matrix. 

The data is simulated according to the model 
$$ y = X\beta + \sigma\epsilon, \epsilon \sim \mathcal{N}(0,1).$$
The data generated in each example consists of a training set, a validation set to tune the parameters, and a 
test set to evaluate the
performance of our chosen model according to the measures described above. Following the notation from \cite{enet}, we denote $././.$ the
number of observations in the training, validation and test sets respectively.

\begin{subsubsection}{Orthogonal components example}
 We generate an example with two connected components, where the predictors in different components are orthogonal. The 
 corresponding sample covariance matrix is block diagonal with 2 blocks. As mentioned earlier, the subgradient equations of the lasso and elastic net split
 naturally when the components are orthogonal. Therefore, the component lasso only differs from the non 
 naive elastic net in the NNLS reweighting step. 
 
 We generate the example as follows: $p=8$, $\sigma=3$ and $\beta=(3,1.5,0,0,2,3,0,0)$. We simulate 100 20/20/200 sets of observations such that the 
 correlations within a component are equal to 0.8, and force the correlations between the components to be exactly 0. We then check the performance of
 the component lasso in two settings: when the number of components it uses is fixed to 2, and, when the optimal number of components is chosen in the  
 validation step. The corresponding test MSEs are given in Table~\ref{table:orthog}.
 
 \begin{table}
\footnotesize
  \begin{center}
      \begin{tabular}{ | l | l | p{2cm}  | p{2cm} |}
      \hline
      \textbf{Method} & \textbf{Median MSE} & \textbf{Median FP} & \textbf{Median FN}\\ \hline
      Lasso & 7.16 (0.50) & 0.40 (0.02) & 0 (0.03)\\ \hline
      Rescaled Lasso & 7.26 (0.49) & 0.33 (0.02) & 0 (0.02) \\ \hline
      Lasso-OLS Hybrid & 7.64 (0.46) & 0.33 (0.02) & 0 (0.02) \\ \hline
      Naive Elastic Net & 6.04 (0.45) & 0.43 (0.01) & 0 (0.02)\\ \hline 
      Elastic Net & 5.8 (0.4) & 0.50 (0.01) & 0 (0.02)\\ \hline 
      Ridge & 6.27 (0.44) & 0.50 (0.01) & 0 (0)\\ \hline
      Component Lasso (2 components) & 5.33 (0.36) & 0.43 (0.01) & 0 (0.02) \\ \hline
      Component Lasso & 4.76 (0.34) & 0.43 (0.01) & 0 (0.01) \\ \hline
       \end{tabular}
      \caption{\em Median MSE, false positive and false negative rates for all regression methods when predictors in different components are
      orthogonal. Numbers in parentheses are the standard
      errors. The component lasso--- with two components, and, when the number of components is chosen at the validation step---
      achieves the lowest MSE.}
      \label{table:orthog}
    \end{center}
  \end{table}
 
 The lower test error achieved by the component lasso indicates that for this simulation, the use of NNLS to weight the predictors within each
 component is more advantageous than rescaling the entire predictor vector at once, as in the non naive elastic net.
 
\end{subsubsection}

\begin{subsubsection}{Further examples}
We consider four examples. The first and third examples are from the original lasso paper \cite{lasso}. The covariance matrix in those examples is not 
block diagonal, 
so the efficiency of the component lasso method in such a setting is not clear apriori. In the second example, we simulate a set-up that seems
well adapted
to the component lasso because the covariance matrix is block diagonal. The variables are split into two connected components. We 
test two instances of this example: one with noise and signal variables in both components, and another with a component containing only 
noise variables. The fourth example is taken from the elastic net paper \cite{enet}. All signal
variables in that example belong to three connected components, and the remaining noise variables are independent. The elastic net
is known to perform well under such conditions, and is shown in \cite{enet} to be better than the lasso at picking out the relevant correlated variables.

Our examples were generated as follows:
\begin{itemize}
 \item Example 1: $p=8$, $\sigma=3$ and $\beta=(3,1.5,0,0,2,0,0,0)$. We simulate 100 20/20/200 sets of observations with pairwise
 correlation ${\rm corr}(i,j)=0.5^{|i-j|}$. This gave an average signal to noise ratio (SNR) of 2.38.
 \item Example 2: $p=8$, $\sigma=5$ and $\beta=(3,1.5,0,0,2,3,0,0)$ or $\beta=(3,1.5,2,3,0,0,0,0)$. We simulate 100 20/20/200 sets of observations
 in the following way:
 $$ x_{i} = Z_1 + \epsilon_i \text{ if } i \in 1,\ldots,4 $$
 $$ x_{i} = Z_2 + \epsilon_i \text{ if } i \in 5,\ldots,8 $$
 where $Z_1$ and $Z_2$ $\sim \mathcal{N}(0,2)$ and $\epsilon_i \sim \mathcal{N}(0,0.5)$. The main point of this example is to compare the performance
 of the component lasso depending 
 on whether the signal variables are in separate connected components (signal in $C_1$ and $C_2$) or in the same one (signal in $C_1$). The 
 respective 
 average SNRs were 4.68 and 8.73.
 \item Example 3: $p=40$, $\sigma=15$ and $$\beta=(\underbrace{0,\dots,0}_{10},\underbrace{2,\dots,2}_{10},\underbrace{0,\dots,0}_{10},
 \underbrace{2,\dots,2}_{10}).$$ We simulate 100 100/100/400 sets of observations with pairwise correlations ${\rm corr}(i,j)=0.5$ if $i\neq j$. 
 This gave an average SNR of 7.72.
 \item Example 4: $p=40$, $\sigma=15$ and $$\beta=(\underbrace{3,\dots,3}_{15},\underbrace{0,\dots,0}_{25}).$$ 
 The predictors are generated according to 3 correlated groups. We simulate 100 50/50/200 sets of observations according to the following model 
 from \cite{enet}:
 $$ x_i=Z_1+\epsilon_i^x, Z_1 \sim \mathcal{N}(0,1), i=1,\dots,5, $$
 $$ x_i=Z_2+\epsilon_i^x, Z_2 \sim \mathcal{N}(0,1), i=6,\dots,10,$$
 $$ x_i=Z_3+\epsilon_i^x, Z_3 \sim \mathcal{N}(0,1), i=11,\dots,15,$$
 where $ \epsilon_i^x \sim \mathcal{N}(0,0.01) \text{ for } i \in 1,\dots,15$ and $x_i \sim \mathcal{N}(0,1) \text{ for } i \in 16,\dots,40$. 
 The corresponding correlations matrix has a block-diagonal structure. This gave an average SNR of 2.97. 
\end{itemize}

Heat maps of sample covariance matrices corresponding to the above examples are shown in Figure~\ref{fig:covs}.
\begin{figure}
\begin{minipage}[t]{0.4\textwidth}
\includegraphics[width=\linewidth]{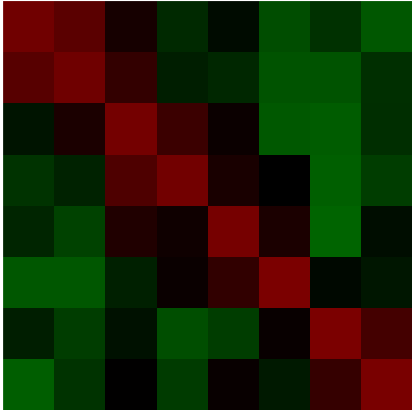}
\end{minipage}
\hspace{\fill}
\begin{minipage}[t]{0.4\textwidth}
\includegraphics[width=\linewidth]{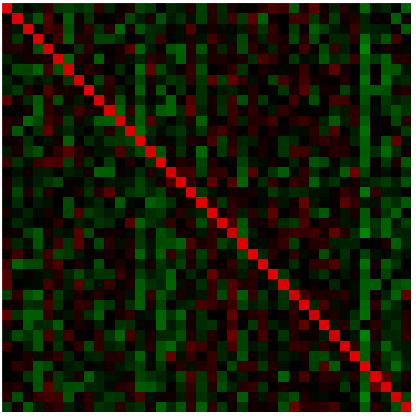}
\end{minipage}
\vspace{1cm}

\begin{minipage}[t]{0.4\textwidth}
\includegraphics[width=\linewidth]{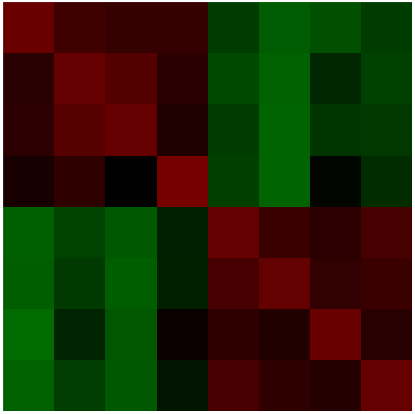}
\end{minipage}
\hspace{\fill}
\begin{minipage}[t]{0.4\textwidth}
\includegraphics[width=\linewidth]{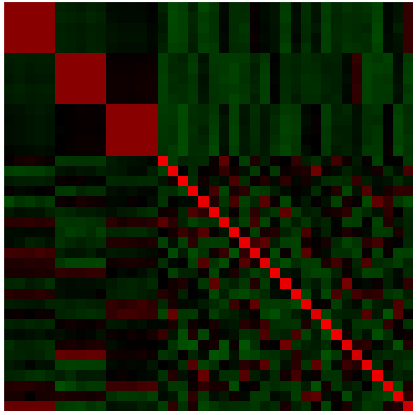}
\end{minipage}
\vspace{1cm}
\caption{\em Heat maps of the sample covariance matrices. Examples 1 (top left) and 3 (top right) do not have a block-diagonal structure, whereas
examples 2 (bottom left) and 4 (bottom right) do.}
\label{fig:covs}
\end{figure}

Table~\ref{table:allsims} shows the results of common penalized regression methods on the above examples: the median MSE, median false positive and false negative rates. 
The component lasso performs well 
in all examples, including the ones where the data is not generated according to a covariance matrix with a block structure. The MSE achieved by the
component
lasso is the lowest. The use of the estimated connected components introduces a more significant improvement in example 2 when the signal variables
are in the same component,
and in example 4 (indicated by a *). The model for both of these datasets has a block-diagonal covariance matrix, where certain components contain
only signal variables, and the
remaining components contain only noise variables. The NNLS reweighting step helps select the components containing the signal predictors.
\begin{table}
\footnotesize
  \begin{center}
      \begin{tabular}{ | l | l | p{2cm}  | p{2cm} |}
      \hline
      \textbf{Method} & \textbf{Median MSE} & \textbf{Median FP} & \textbf{Median FN}\\ \hline
      \textbf{Example 1} & & & \\ \hline
      Lasso & 2.44 (0.28) & 0.50 (0.02) & 0 (0.02)\\ \hline
      Rescaled Lasso & 2.16 (0.26) & 0.40 (0.02) & 0 (0.02) \\ \hline
      Lasso-OLS Hybrid & 2.10 (0.25) & 0.25 (0.02) & 0 (0.01) \\ \hline
      Naive Elastic Net & 2.17 (0.26) & 0.50 (0.02) & 0 (0.02)\\ \hline 
      Elastic Net & 1.82 (0.25) & 0.50 (0.02) & 0 (0.02)\\ \hline 
      Ridge & 2.79 (0.28) & 0.62 (0) & 0 (0)\\ \hline
      Component Lasso & 1.59 (0.22) & 0.40 (0.02) & 0 (0.02) \\ \hline
      \textbf{Example 2 (Signal in $C_1$ and $C_2$)} & & & \\ \hline
      Lasso & 7.63 (0.55) & 0.37 (0.02) & 0 (0.02)\\ \hline
      Rescaled Lasso & 7.17 (0.58) & 0.33 (0.02) & 0 (0.02) \\ \hline
      Lasso-OLS Hybrid& 7.48 (0.58) & 0.25 (0.02) & 0.2 (0.02) \\ \hline
      Naive Elastic Net & 6.08 (0.48) & 0.43 (0.01) & 0 (0.02)\\ \hline 
      Elastic Net & 5.87 (0.40) & 0.50 (0.01) & 0 (0.02)\\ \hline 
      Ridge & 6.61 (0.47) & 0.50  (0) & 0    (0)\\ \hline
      Component Lasso & 4.89 (0.33 )& 0.43 (0.01)  & 0 (0.02) \\ \hline
      \textbf{Example 2 (Signal in $C_1$)} & & & \\ \hline
      Lasso & 5.95 (0.53) & 0.25 (0.02) & 0 (0.02)\\ \hline
      Rescaled Lasso & 5.49 (0.44) & 0.20 (0.02) & 0 (0.02)\\ \hline
      Lasso-OLS Hybrid& 5.31 (0.47) & 0 (0.01) & 0.2 (0.01) \\ \hline
      Naive Elastic Net & 4.14 (0.47) & 0.33 (0.01) & 0 (0.01)\\ \hline 
      Elastic Net & 1.83 (0.27) & 0 (0.02) & 0 (0)\\ \hline 
      Ridge & 4.4 (0.5) & 0.50  (0) & 0  (0)\\ \hline
      Component Lasso & 1.57* (0.27) & 0 (0.02) & 0 (0) \\ \hline
      \textbf{Example 3} & & & \\ \hline
      Lasso & 58.61 (1.43) & 0.31 (0.01) & 0.23 (0.01)\\ \hline
      Rescaled Lasso & 58.44 (1.54) & 0.31 (0.01) & 0.25 (0.01)  \\ \hline
      Lasso-OLS Hybrid & 57.25 (1.64) & 0.28 (0.01) & 0.24 (0.01) \\ \hline
      Naive Elastic Net & 38.74 (0.93) & 0.41 (0) & 0.14 (0.01)\\ \hline 
      Elastic Net & 31.75 (0.69) & 0.46 (0) & 0 (0.02)\\ \hline 
      Ridge & 32.86 (0.74) & 0.50 (0) & 0  (0)\\ \hline
      Component Lasso & 31.16 (0.73) & 0.46 (0) & 0 (0.02)\\ \hline
      \textbf{Example 4} & & & \\ \hline
      Lasso & 46.62 (3.29) & 0.60 (0.01) & 0.37 (0.01)\\ \hline
      Rescaled Lasso & 28.67 (3.09) & 0.29 (0.02) & 0.31 (0) \\ \hline
      Lasso-OLS Hybrid & 15.75 (2.02)  & 0 (0.01) & 0.32 (0.02)\\ \hline
      Naive Elastic Net &  44.90 (3.01) & 0.47 (0.01) & 0.20 (0.01)\\ \hline 
      Elastic Net & 23.79 (2.66) & 0.25 (0.03) & 0 (0.01)\\ \hline 
      Ridge & 61.74 (3.99) & 0.62   (0) & 0    (0) \\ \hline
      Component Lasso & 10.74* (2.34) & 0.06 (0.01) & 0.04 (0.01) \\ \hline
      \end{tabular}
      \caption{\em  Median MSE, false positive and false negative rates for the four simulated examples
      using 7 regression methods. Numbers in parentheses are the standard errors.  }
    \label{table:allsims}
   \end{center}
  \end{table}

  For every data set, the connected-component split which gave the lowest validation MSE is chosen to compute the test error. Tables 2-6 show
  the distribution of the number of components that minimize the error in all examples. The number of components (NOC) by itself is not an appropriate measure 
  to verify how the
  predictors are being grouped. {For example, consider the case where some of the connected components only contain noise variables. Then, whether
  those variables are grouped correctly or kept in one big component does not affect the performance of the component lasso as long as the noisy components
  are excluded.} In order to focus
  on how the signal variables are split, we use the misclassification measure from \citeasnoun{CT2005} on the signal variables only: 
   $$M(C,T)=\frac{ \sum_{i>i'}|I_C(i,i')-I_T(i,i')| }{{n \choose 2}} , $$
  where C is the partition of points, T corresponds to the true clustering, and $I()$ is an indicator function for 
  whether the clustering places i and i' in the same cluster. The measure quantifies the misclassification of signal variables over all signal pairs. It can
  be seen from the tables that the component lasso method favors splitting the predictors into clusters with low misclassification
  rate. The true number of components, which corresponds to the number of diagonal blocks in the covariance matrix used to generate the data, is
  indicated by a *.
  
  \begin{table}[h!]
  \small
   \begin{center}
      \begin{tabular}{ | l | l | l| l| l |}
      \hline
       \textbf{Number of Components} & 1* & 3 & 5 & 7  \\ \hline
       \textbf{Number of Datasets } & 38 & 26 & 21 & 15 \\ \hline
       \textbf{Mis. Rate } & 0 & 0.60 & 0.86 & 1  \\ \hline
        \end{tabular}
        \caption{\em  \textbf{Example 1:} Optimal NOC and misclassification rate of the signal variables. }
   \end{center}
  \end{table}
  
  \begin{table}[h!]
  \small
   \begin{center}
      \begin{tabular}{ | l | l | l| l| l | l | l | l | l| }
      \hline
       \textbf{N. of Components} & 1 & 2* & 3 & 4 & 5 & 6 & 7 & 8   \\ \hline
       \textbf{N. of Datasets } & 35 & 14 & 6 & 6 & 9 & 12 & 12 & 6  \\ \hline
       \textbf{Mis. Rate } & 0.67 & 0 & 0.17 & 0.20 & 0.18 & 0.25 & 0.28 & 0.33 \\ \hline
        \end{tabular}
        \caption{\em \textbf{Example 2 (Signal in $C_1$ and $C_2$):} Optimal NOC  and misclassification rate of the signal variables. }
   \end{center}
  \end{table}
  
  \begin{table}[h!]
  \small
   \begin{center}
      \begin{tabular}{ | l | l | l| l| l | l | l | l | l| }
      \hline
       \textbf{N. of Components} & 1 & 2* & 3 & 4 & 5 & 6 & 7 & 8   \\ \hline
       \textbf{N. of Datasets } & 45 & 11 & 16 & 8 & 10 & 8 & 2 & 0  \\ \hline
       \textbf{Mis. Rate } & 0 & 0 & 0.43 & 0.63 & 0.55 & 0.71 & 0.92 & - \\ \hline
        \end{tabular}
        \caption{\em \textbf{Example 2 (Signal in $C_1$):} Optimal NOC and misclassification rate of the signal variables. }
   \end{center}
  \end{table}

  \begin{table}[h!]
  \small
   \begin{center}
      \begin{tabular}{ | l | l | l| l| l | l|  l | l| l| l| }
      \hline
       \textbf{N. of Components} & 1* & 5 & 9 & 13 & 17 & 21 & 25   \\ \hline
       \textbf{N. of Datasets } & 59 & 18 & 13 & 5 & 3 & 1 & 1 \\ \hline
       \textbf{Mis. Rate } & 0 & 0.32 & 0.65 & 0.88 & 0.89 & 0.98 & 0.95 \\ \hline
        \end{tabular}
        \caption{\em \textbf{Example 3:} Optimal NOC and misclassification rate of the signal variables. }
        \end{center}
  \end{table}

  \begin{table}[h!]
  \small
   \begin{center}
      \begin{tabular}{ | l | l | l| l| l | l| l | l | l | l | l |}
      \hline
       \textbf{N. of Components} & 1 & 5 & 9 & 13 & 17 & 21 & 25 & 29* & 33 & 37  \\ \hline
        \textbf{N. of Datasets } & 20 & 28 & 14 & 2 & 3 & 1 & 2 & 15 & 11 & 4  \\ \hline
        \textbf{Mis. Rate } & 0.71 & 0.07 & 0.03 & 0 & 0 & 0 & 0 & 0.04 & 0.16 & 0.25 \\ \hline
        \end{tabular}
        \caption{\em \textbf{Example 4:} Optimal NOC and misclassification rate of the signal variables. The true NOC is 28. 29 is the closest value in
        the tested grid. }
   \end{center}
  \end{table}
  \end{subsubsection}
 
\subsection{Real data example}

The component lasso is designed for settings where the data consist of a large number of predictors which can be split into highly correlated subgroups.
We use a dataset from genetics to evaluate the performance of the method, because data in this area tend to follow this structure.
Molecular markers are fragments of DNA associated with certain locations in the genome. In recent years, the
abundance of molecular markers has made it possible to use them to predict genetic traits using linear regression.
The genetic value of genes that influence a trait of interest is defined as the average phenotypic value over individuals with that trait. A standard
genetic model consists in writing 
the phenotype $y$ as a sum of genetic values such that $y=X\beta+\epsilon$, where $X$ contains genetic values of the considered molecular markers. \\

Here, we consider the wheat data set studied in~\cite{GYp2010}. The aim is to predict genetic values of a quantitative trait,
specifically grain yield in a fixed type of environment. The dataset consists of 599 observations, each corresponding to a different wheat line. Following
the analysis done in~\cite{GYp2010}, we use 1279 predictors which indicate the presence or absence of molecular markers.
The grain yield response is available in 4 distinct environments. We normalize the data so that 
the predictors are centered and scaled, and the response is centered. We then split the available observations into equally sized training and test sets.
Finally, we apply cross validation to determine the model parameters. \\

The test MSE is defined as $\sum_i(y_i-\hat{y}_i)^2/n$ for $i$ in the test set. We compare the error rates of the lasso, naive elastic net, elastic net 
and the component lasso. We fix the range of the number of components for the component lasso to be between 1 and 50. 
Table~\ref{table:real} contains the test MSE achieved by the different methods to predict grain yield in 4 environments. \\
 
\begin{table}[h]
\begin{center}
      \begin{tabular}{ | l | l | l | l |}
      \hline
      \textbf{Method} & \textbf{Test MSE} & \textbf{Parameters} & \textbf{Variables Selected} \\ \hline
      \textbf{Environment 1} & & & \\ \hline
      Lasso & 0.8547 & $\lambda=5.1e^{-2}$ & 45 \ \\ \hline
      Naive Elastic Net &  0.9656 & $\alpha=0.05$, $\lambda=2.43e^{-4}$ & 1195 \\ \hline 
      Elastic Net & 0.9122 & $\alpha=0.05$, $\lambda=9.58e^{-4}$ & 1151 \\ \hline 
      Component Lasso & 0.7552 & $\alpha=0.05$, $\lambda=1.8e^{-4}$, $noc=29$ & 548  \\ \hline
      \textbf{Environment 2}  & & & \\ \hline
      Lasso & 0.8875 & $\lambda=6.18e^{-3}$ & 38 \ \\ \hline
      Naive Elastic Net &  1.1104 & $\alpha=0.05$, $\lambda=2.68e^{-4}$ & 1191 \\ \hline 
      Elastic Net & 1.0722 & $\alpha=0.05$, $\lambda=6.57e^{-4}$ & 1170 \\ \hline 
      Component Lasso & 0.8775 & $\alpha=0.05$, $\lambda=2.45e^{-5}$, $noc=37$ & 564  \\ \hline
      \textbf{Environment 3}  & & & \\ \hline
      Lasso & 0.8216 & $\lambda=7.23e^{-3}$ & 28 \ \\ \hline
      Naive Elastic Net &  1.1087 & $\alpha=0.05$, $\lambda=4.89e^{-4}$ & 1170 \\ \hline 
      Elastic Net & 1.1249 & $\alpha=0.05$, $\lambda=1.20e^{-3}$ & 1125 \\ \hline 
      Component Lasso & 0.8830 & $\alpha=0.05$, $\lambda=1.62e^{-3}$, $noc=17$ & 303  \\ \hline
      \textbf{Environment 4} & & & \\ \hline
      Lasso & 0.8068 & $\lambda=6.8e^{-3}$ & 27 \ \\ \hline
      Naive Elastic Net &  1.0487 & $\alpha=0.05$, $\lambda=5.18e^{-4}$ & 1157 \\ \hline 
      Elastic Net & 0.9349 & $\alpha=0.05$, $\lambda=2.75e^{-3}$ & 1081 \\ \hline 
      Component Lasso & 0.8200 & $\alpha=0.05$, $\lambda=3.36e^{-5}$, $noc=33$ & 564  \\ \hline
      \end{tabular}
      \caption{\em Test MSE, parameters, and number of non-zero predictors for the real data set. }
      \label{table:real}
 \end{center}
\end{table}

\begin{figure}[!h]
\centering
  \includegraphics[scale=0.4]{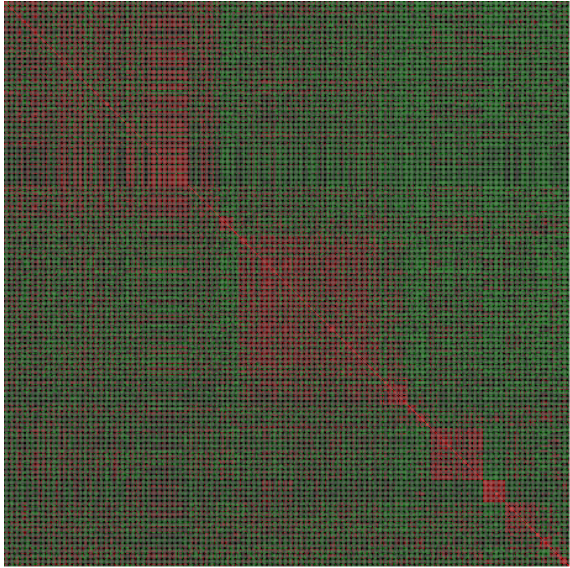}
  \caption{\em Heat map of the sample covariance matrix corresponding to the wheat data.}
  \label{fig:realheat}
\end{figure}

The component lasso achieves the lowest test MSE in environments 1 and 2 by splitting the variables into 29 and 37 connected components respectively. The 
lasso achieves the lowest MSE in environments 3 and 4. In this dataset, splitting the genomic markers into correlated groups helped improve the accuracy in
certain environments. Sorting the predictors according to the connected components chosen by the component lasso and plotting the heat map of the sample 
covariance matrix reveals the block-diagonal structure of the genomic markers. The corresponding connected components can be seen in
Figure~\ref{fig:realheat}.

This example illustrates that the component lasso 
can provide improved prediction accuracy and interpretability in some real data problems.

\subsection{Recovery of the true non-zero parameter support}

\label{LM.sec.further.lasso}
There has been much study of the ability of the lasso and related
procedures to recover the correct model, as $n$ and $p$ grow.
Examples of this work include \citeasnoun{KF2000},
\citeasnoun{GR2004}, \citeasnoun{tropp2004}, \citeasnoun{donoho2006},
\citeasnoun{Mein2007}, \citeasnoun{MB2006}, \citeasnoun{tropp2006},
\citeasnoun {ZY2006}, \citeasnoun{wainwright2006}, and
\citeasnoun{BTW2007}. 

Many of the results in this area assume an ``irrepresentability'' condition on the design
matrix of the form
\begin{eqnarray}
||({X_\cS}^TX_\cS)^{-1}{X_{\cS}}^TX_{\cS^c}{\rm sign}(\beta_1)||_\infty \leq (1-\epsilon)\; \mbox{for
  some}\; \epsilon \in (0,1]
\label{LM.lassocond}
\end{eqnarray}
\cite{ZY2006}.
The set $\cS$ indexes the subset of features with non-zero coefficients in the true
underlying model, and $X_\cS$ are the columns of $X$ corresponding
to those features. Similarly  $\cS^c$ are the features with
true coefficients equal to zero, and $X_{\cS^c}$ the corresponding
columns. The vector $\beta_1$ denotes the coefficients of the non-zero
signal variables.
The condition~\eqref{LM.lassocond} says that the least squares coefficients for the columns of  $X_{\cS^c}$ on
$X_\cS$ are not too large, that is, the ``good'' variables $\cS$ are
not
too highly correlated with the nuisance variables $\cS^c$.

Now suppose that the signal variables and noise variables
fall into two separate components $C_1, C_2$ with sufficient within-component
correlation that we are able to identify them from the data.
Note that $C_1$ might also contain some noise variables.
Then in order to recover the signal successfully, we need only that the
noise variables within $C_1$ are irrepresentable by the signal variables,
as opposed to all noise variables. This result follows from the fact that for
block diagonal correlation matrices, the strong irrepresentable condition holds if
and only there exists a common $0<\eta \leq 1$ for which the strong irrepresentable condition
holds for every block.

\subsection{Grouping effect}

The grouping effect refers to the property of a regression method that returns similar coefficients for highly correlated variables. If some 
predictors happen to be identical, the method should return equal coefficients for the corresponding variables. The  elastic net is shown to exhibit this
property in the extreme case where predictors are identical ( \cite{enet} lemma 2). Moreover, in Theorem 1 of the same paper, the authors bound
the absolute value of the difference between
coefficients $\hat{\beta}_i$ and $\hat{\beta}_j$ in terms of their sample correlation $\rho=x_i^Tx_j$.

In the component lasso method, we use the elastic net or the lasso to estimate the coefficients of every connected component. If we assume
that we are able to 
identify the components correctly from the data, then the first step of the component lasso method will
preserve the grouping effect when the elastic net is used for every subproblem. NNLS fitting will also preserve the property since variables within the same connected component
are scaled by the same coefficient.

\section{Computational Considerations}
\label{sec:comp}
We use the {\tt glmnet} 
package in R for fitting the lasso and elastic  net
\cite{friedman08:_regul_paths_gener_linear_model_coord_descen}.
This package uses cyclical coordinate descent 
using a ``naive'' method for $p>500$ and a ``covariance'' mode
for $p \le 500$.
Empirically, the computation time for the algorithm in naive mode scales as $O(np^2)$ (or perhaps $O(np^{1.5})$).

Now suppose we divide the predictors into $L$ connected components:
this requires $O(np^2)$ operations and can  be done without forming {the sample covariance matrix} $S$
(see  e.g. \citeasnoun{M2002}).
The lasso or elastic net fitting in each of the components takes $O(Ln (p/L)^2)=O(np^2/L)$.
The final non-negative least squares fit can be done in $O(nL^2)$.
Thus the overall computational complexity of the component lasso is
about the same as for the lasso itself.

Table~\ref{tab:timings} shows some sample timings for  agglommerative  clustering with different
linkage methods applied
to problems with different $n$ and $p$. The $p$ columns of $X$ were 
clustered and the code ran on a standard linux server.
We used the {\tt Rclusterpp} R package from the CRAN repository.
\begin{table}
\begin{center}
\begin{tabular}{|rr|rrrr|r|}
\hline
&&\multicolumn{4}{c}{Linkage} & glmnet\\ 
   n &   p  &    ave &   comp &  sing &  Ward& \\
\hline
200&200&0.372&0.172&0.020&0.024&0.312\\
200&1000&20.698&2.852&0.268&0.412&0.056\\
200&2000&103.382&11.225&1.040&1.792&0.076\\
1000&200&1.816&0.572&0.048&0.080&0.036\\
2000&200&3.484&1.108&0.100&0.188&0.056\\
2000&1000&191.468&27.574&3.200&8.409&0.960\\
2000&2000&1316.306&121.244&14.765&42.722&9.453\\
\hline
\end{tabular}
\end{center}
\caption[tab:timings]{\em Timings for various hierarchical clustering techniques,
compared to {\em glmnet}}
\label{tab:timings}
\end{table}

We see that columns scale roughly as $O(np^2)$,
but some linkages are much faster than others.
However the computational  time for the lasso fit by {\tt glmnet}
seems to grow more slowly than that for the clustering operations.

However, there is potential for significant speedups in the component lasso 
algorithm.
The main bottleneck is the clustering step, which requires about $O(np^2)$ operations, as seen above.
But in fact we do not need to cluster all $p$ features.
If a feature is never entered into the model, we don't need to determine its cluster membership
and hence don't need to compute its inner product  with other features.

Consider for example the covariance mode of  {\tt glmnet}.
 Suppose we have a model with $k$ nonzero coefficients.
 For {\tt glmnet}, we need to  compute the inner products of these features with all other features, $kp$ in all.
For the component lasso, suppose that we have $K$ clusters of equal size, and $k/K$ nonzero coefficients in each.
Then we only need to compute  $K(k/K)(p/K)=kp/K$ inner products, plus the number  needed
to determine the cluster memberships of
clusters containing each of the $k$ features. 
This is $O(p)$ inner products.
Thus the total number is reduced from $O(kp)$ to $O(kp/K+ p)$.
A careful implementation of this procedure will be done in future work. 

We note that  cross validation is potentially slower for the component lasso
since it needs to consider splitting the covariance matrix into multiple numbers of components.
This results in an extra parameter--- the number of components---
that must be varied in the cross-validation step.

Finally, the
the modular structure of the component lasso lends itself naturally to parallel computation.
This will also be developed in future work.

\section{Discussion}
\label{sec:discussion}

In this paper we have proposed the component lasso, a penalized  regression and variable selection method. In particular, we have shown that estimating and exploiting  the
block-diagonal structure of the sample covariance matrix--- solving separate lasso problems and then recombining--- can yield more
accurate predictions and better recovery of the support of the signal.
 We provide simulated and real data examples where the component lasso outperforms standard 
regression methods in terms of prediction 
error and support recovery.

There are possible extensions of this work to other settings.
Consider a $\ell_1$-penalized logistic regression model with outcome $y_i \in [0,1]$,
 $\mu={\rm Pr}(Y=1|x)$ and linear predictor $\eta=\log(\mu/(1-\mu))=\beta_0+x\beta$.
Then the subgradient equations have the form
\begin{equation}
  X^T W X\beta - X^T Wz + \lambda\cdot \text{sign}(\beta) = 0 
\label{eqn:irls}
\end{equation}
with $z=\beta_0+X\beta$ and $W={\rm diag}(\mu_1, \mu_2,\ldots  \mu_n)$.
Typical algorithms start with some initial value $\beta'$, compute
$W$ and $z$ and then solve (\ref{eqn:irls}).
Then $W$ and $z$ are updated and the process is repeated until convergence.
This is known as iteratively reweighted (penalized)  least squares (IRLS).

We see that the appropriate connected components
are those of $X^T WX$: however this depends on $\beta$ and would have to be
re-computed at each iteration. We might instead set $\beta'=0$ so that
$X^T WX= X^TX/4$.
Hence we find the connected components of $ X^TX$
and fix them. This leads to $K$ separate $\ell_1$-penalized
logistic regression problems with estimates $\hat\eta_1, \hat\eta_2, \ldots \hat\eta_K$. These could be combined by a non-negative-constrained logistic regression of 
$y$ on
$\{\hat\eta_\ell, k=1,2,\ldots K\}$.
An analogous approach could be used for other generalized linear models.


The component lasso achieves a significant reduction in prediction error in examples for which the covariance matrix has a block-diagonal structure
and where some
components only contain noise variables. The NNLS step allows the component lasso to select the relevant components due to the fact that it
induces 
sparsity in the estimated coefficients. The component lasso also exhibits a better performance in other examples, in which NNLS helps by weighting 
the contribution of each component. Thus the properties of NNLS are crucial to the performance of the method. In future work
we will study the theoretical properties of the component lasso.

\subsection*{Acknowledgements}
The authors thank Trevor Hastie for helpful suggestions.
Robert Tibshirani was supported by National Science Foundation Grant DMS-9971405 and
National Institutes of Health Contract N01-HV-28183.



%
%
%
\bibliographystyle{agsm}
\bibliography{tibs}

\end{document}